\documentclass{article}
\usepackage[utf8]{inputenc} 
\usepackage[T1]{fontenc}    
\usepackage{microtype}      
\usepackage{nicefrac}       
\usepackage[colorlinks = true,
            hypertexnames = false,
            linkcolor = black,
            urlcolor  = blue,
            citecolor = blue,
            anchorcolor = blue]{hyperref}
\usepackage{url}            
\usepackage{booktabs}       
\usepackage{amsfonts}       
\usepackage{xcolor}         
\usepackage{amssymb, amsmath, amsthm, amstext}
\usepackage{verbatim}
\usepackage{bbm}
\usepackage{enumerate}
\usepackage{dsfont}
\usepackage[mathscr]{eucal}
\usepackage{graphicx}
\usepackage{mathrsfs}
\usepackage[ruled,vlined]{algorithm2e}
\usepackage{wrapfig}
\usepackage[nameinlink]{cleveref} 
\usepackage{tikz}
\usepackage{placeins}
\usepackage{subfigure}
\usepackage{bm}
\usepackage[preprint]{StylingInput}

\title{dUltra: Ultra-Fast Diffusion Language Models via Reinforcement Learning}
\author{Shirui Chen \\
University of Washington\\
\texttt{sc256@uw.edu} \\
\And
Jiantao Jiao \\
University of California, Berkeley\\
\texttt{jiantao@eecs.berkeley.edu} \\
\AND
Lillian J. Ratliff \\
University of Washington\\
\texttt{ratliffl@uw.edu} \\
\And
Banghua Zhu \\
University of Washington\\
\texttt{banghua@uw.edu}
}
\date{\today{}}
\begin{document}
\maketitle

\begin{abstract}
Masked diffusion language models (MDLMs) offer the potential for parallel token generation, but most open-source MDLMs decode fewer than 5 tokens per model forward pass even with sophisticated sampling strategies, limiting their parallel generation potential. Existing acceleration methods either rely on fixed confidence-based heuristics or use distillation-based approaches that finetune MDLMs on trajectories generated by a base model, which can become off-policy during finetuning and restrict performance to the quality of the base model's samples.
We propose \texttt{dUltra}, an on-policy reinforcement learning framework based on Group Relative Policy Optimization (GRPO) that learns unmasking strategies for efficient parallel decoding. dUltra introduces an unmasking planner head that predicts per-token unmasking likelihoods under independent Bernoulli distributions. We jointly optimize the base diffusion LLM and the unmasking order planner using reward signals combining verifiable reward, distillation reward, and the number of unmasking steps. Across mathematical reasoning and code generation tasks, dUltra achieves superior accuracy-efficiency trade-offs compared to state-of-the-art heuristic (Fast-dLLM) and distillation baselines (d3LLM, dParallel), demonstrating that learned unmasking trajectories through on-policy RL enable better exploitation of parallel generation in MDLMs. Code and checkpoints are released \href{https://github.com/chinsengi/dUltra-os}{here}.
\end{abstract}

\section{Introduction}

The great success of diffusion models and their parallel generation nature makes them attractive for language generation. There has been an ongoing effort to transfer the diffusion paradigm to the natural language domain. Earlier attempts tried to sample in the token embedding space \citep{NEURIPS2022_1be5bc25,gulrajani2023likelihoodbaseddiffusionlanguagemodels,gong2023diffuseqsequencesequencetext} or the latent space of a language autoencoder \citep{lovelace2023latentdiffusionlanguagegeneration} using continuous diffusion. In a parallel line of work, discrete diffusion in the vocabulary space is studied in \citep{austin2023structureddenoisingdiffusionmodels,campbell2022continuous,meng2022concrete,he2022diffusionbertimprovinggenerativemasked, wang2019bertmouthspeakbert}. Without the need for an invertible token-wise embedding function, discrete diffusion language models generally perform better than continuous diffusion language models when operating in a discrete state space with a countable number of states \citep{lou_discrete_2024,shi_simplified_2024,sahoo_simple_2024}. Moreover, it is observed that the best performance is achieved with an absorbing source distribution, which further restricts possible sampling paths \citep{lou_discrete_2024,austin2023structureddenoisingdiffusionmodels}. The diffusion models with an absorbing source distribution are also called masked diffusion language models (MDLMs) because of the introduction of a masked token. Moreover, \citet{ou2025absorbingdiscretediffusionsecretly} and \citet{sahoo_simple_2024} have shown that the optimal solution that minimizes negative ELBO is independent of the time variable, thus promoting time-agnostic parameterization. Recent works have scaled time-agnostic MDLMs to show that their performance can rival that of autoregressive (AR) models with similar parameter size \citep{nie2025largelanguagediffusionmodels,dream2025}. 

However, even these best open-source masked diffusion language models (MDLMs) \citep{nie2025largelanguagediffusionmodels,dream2025} face the same slow sampling problem as continuous diffusion models. Apart from system-level optimization, such as approximate KV Cache \citep{wu2025fastdllmtrainingfreeaccelerationdiffusion,hu2025acceleratingdiffusionlanguagemodel}, attempts to accelerate MDLMs have focused on designing confidence/entropy-aware sampling methods to unmask tokens in parallel \citep{ben-hamu_accelerated_2025,wu2025fastdllmtrainingfreeaccelerationdiffusion}. These deterministic parallel decoding strategies generally favor tokens with high confidence or low entropy within a single unmasking step, leading to 3-5x faster sampling speed. Despite this decent speed-up, the number of tokens decoded in parallel within a single denoising step is still low ($\sim$4.5 tokens per step on GSM8K and $\sim$3.3 tokens per step on MATH500) \citep{wu2025fastdllmtrainingfreeaccelerationdiffusion}. Meanwhile, commercial diffusion language models \citep{gemini-diffusion, labs2025mercury} operate at a throughput of $\sim$1000 tokens/s, significantly surpassing the state-of-the-art throughput of autoregressive models, which typically range from 100-300 tokens/s \citep{labs2025mercury}. Therefore, current open-source MDLMs have yet to exploit the parallel generation potential of diffusion language models to their fullest and have not achieved "diffusion supremacy" (analogous to quantum supremacy) over autoregressive models.

\begin{figure}[!htb]
   \centering
   \includegraphics[width=0.9\textwidth]{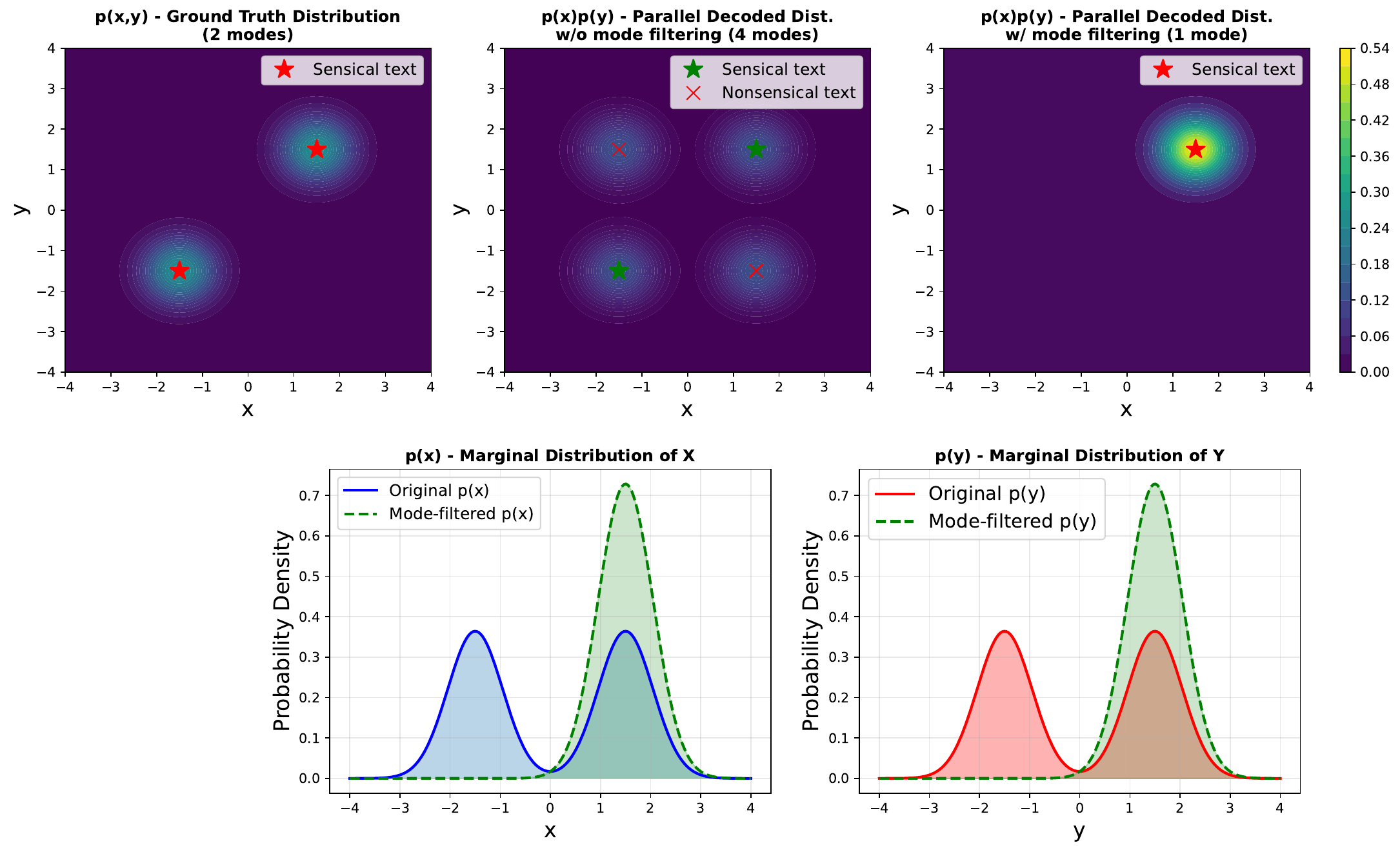}
   \caption{Parallel decoding of dependent ambiguous tokens can lead to unwanted modes.}
   \label{fig:bimodal}
\end{figure}

A fundamental observation is that when tokens are generated in parallel, MDLMs assume conditional independence between them given the masked input sequence (see also \Cref{sec:seq}) due to the curse of dimensionality \citep{dllm_ratio_blog}. If the tokens are actually dependent and ambiguous\footnote{An ambiguous token means that multiple words in the vocabulary are predicted to have similar and high probability.}, the actual distribution sampled from will have unwanted modes, and the model may generate nonsensical text (\Cref{fig:bimodal}, first row, middle panel). Since acceleration methods aim to increase parallelism by decoding more tokens per step, they inevitably unmask some dependent tokens together, violating the independence assumption. Therefore, given the current paradigm of parallel decoding\footnote{Unless curse of dimensionality is solved}, any method that accelerates sampling is implementing some form of mode filtering, i.e. the accelerated sampling procedure will try to sample from a single mode that produces sensical text out of all modes that a slower model can sample from. \Cref{fig:bimodal} illustrates this behavior. 

Recent work \citep{d3llm, chen2025dparallellearnableparalleldecoding} on accelerating diffusion LLM sampling with offline distillation demonstrates this principle as well. Specifically, d3LLM \citep{d3llm} distills from the sampling trajectory produced by the teacher model, effectively teaching the student model to sample from one of the high probability teacher modes at each sampling step. Similarly, dParallel \citep{chen2025dparallellearnableparalleldecoding} proposes certainty-forcing distillation that finetunes the model to be more confident on self-generated trajectories, which are fixed per prompt. This effectively encourages the model to sample from one of the high probability teacher modes marginalized by the prompts. However, this self-distillation approach faces two shortcomings. First of all, the quality of training data cannot exceed the capability of the base model. In other words, the training data only includes greedily sampled trajectories, limiting exploration and performance of the distilled model. Secondly, the self-generated trajectory can be off-policy during distillation, and the student learns in contexts more frequently encountered by the teacher model rather than by itself. Therefore, we want a method that explores the teacher model modes and trains on samples generated by the student model. On-policy reinforcement learning naturally addresses both limitations: by training on trajectories generated by the current policy, it avoids the distribution mismatch inherent in offline distillation, while the exploration encouraged by RL allows discovering better unmasking strategies beyond those demonstrated by the base model.

\begin{figure}[!htb]
   \centering
   \includegraphics[width=\textwidth]{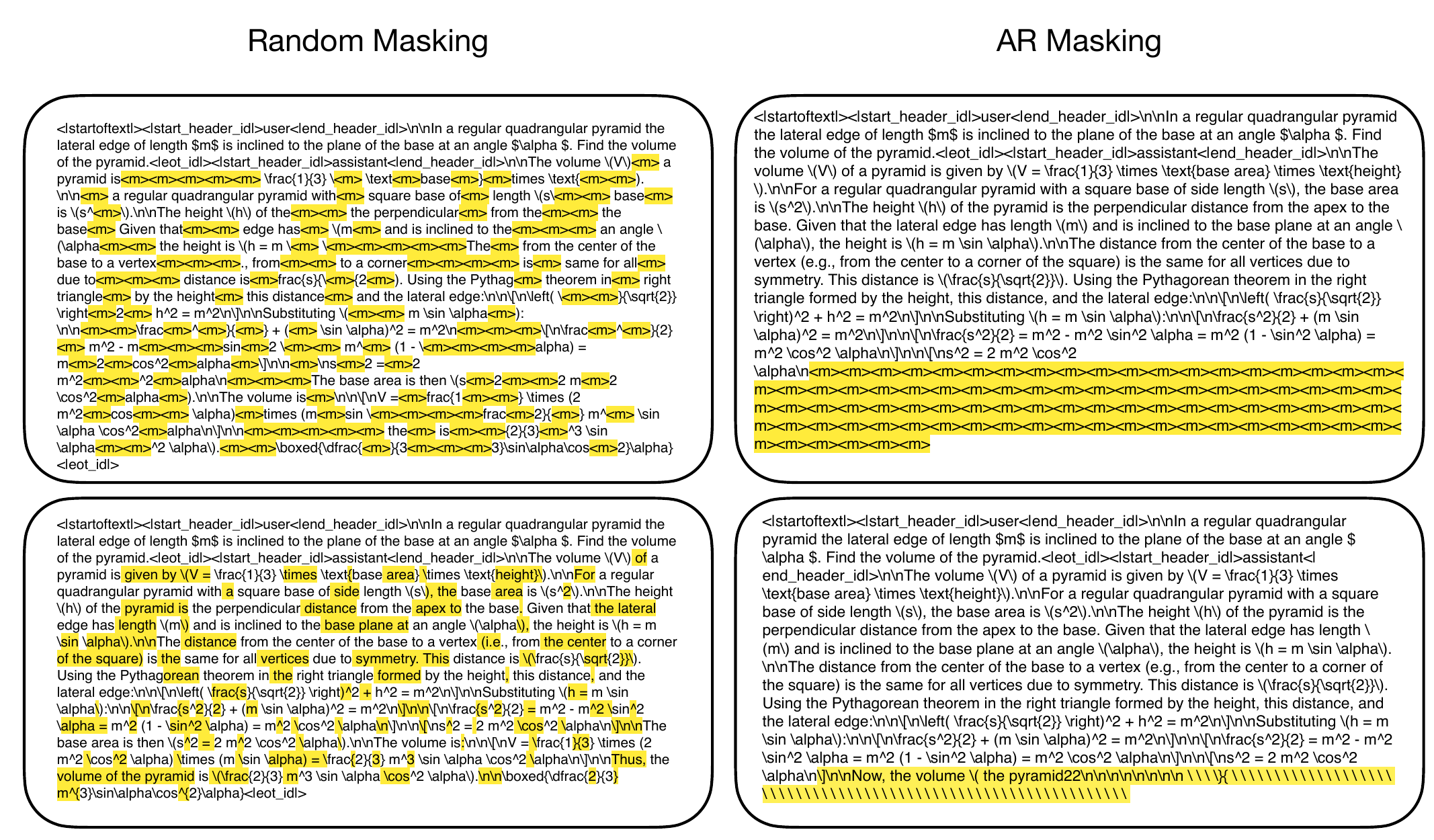}
   \caption{Given the same prompt and ground truth, we mask 30\% of tokens using both random masking and autoregressive (AR) masking. The left column shows the masked input sequence with random masking (\texttt{<m>} represents the mask token), and the corresponding one-step decoding result. The right columns show the masked input sequence with AR masking, and the corresponding one-step decoding result. We use \texttt{LLaDA-8B-Instruct} for both generations.}
   \label{fig:case_study}
\end{figure}

Another implication of the observation above is that we can accelerate sampling while avoiding the need for mode filtering by planning a sampling trajectory such that there are more conditionally independent tokens to decode in each sampling step. To illustrate this intuition, we present a case study comparing the quality of one-step generation of LLaDA-Instruct under different masking strategies in \Cref{fig:case_study}. More specifically, we compare random masking, where we randomly select a certain percentage of tokens in the ground truth to mask, and autoregressive masking, where we mask the same percentage of tokens only at the end of the ground truth sequence. We observe that LLaDA-Instruct is able to decode a large percentage (30\%) of mask tokens with no noticeable quality degradation with random masking, while masking the same percentage of mask tokens at the end of the sequence leads to poor generation quality. This is expected as randomly masked tokens usually have more contextual information making them less conditionally dependent on other masked tokens, while consecutive tokens at the end of the sequence are highly dependent on each other, even if given the partial answer. This case study illustrates the potential of conditional independence-aware planning: our approach will learn adaptive unmasking strategies throughout the full multi-step denoising trajectory to maximize the number of conditionally independent tokens decoded at each step. 

\begin{figure}[!htb]
   \centering
   \includegraphics[width=.9\textwidth]{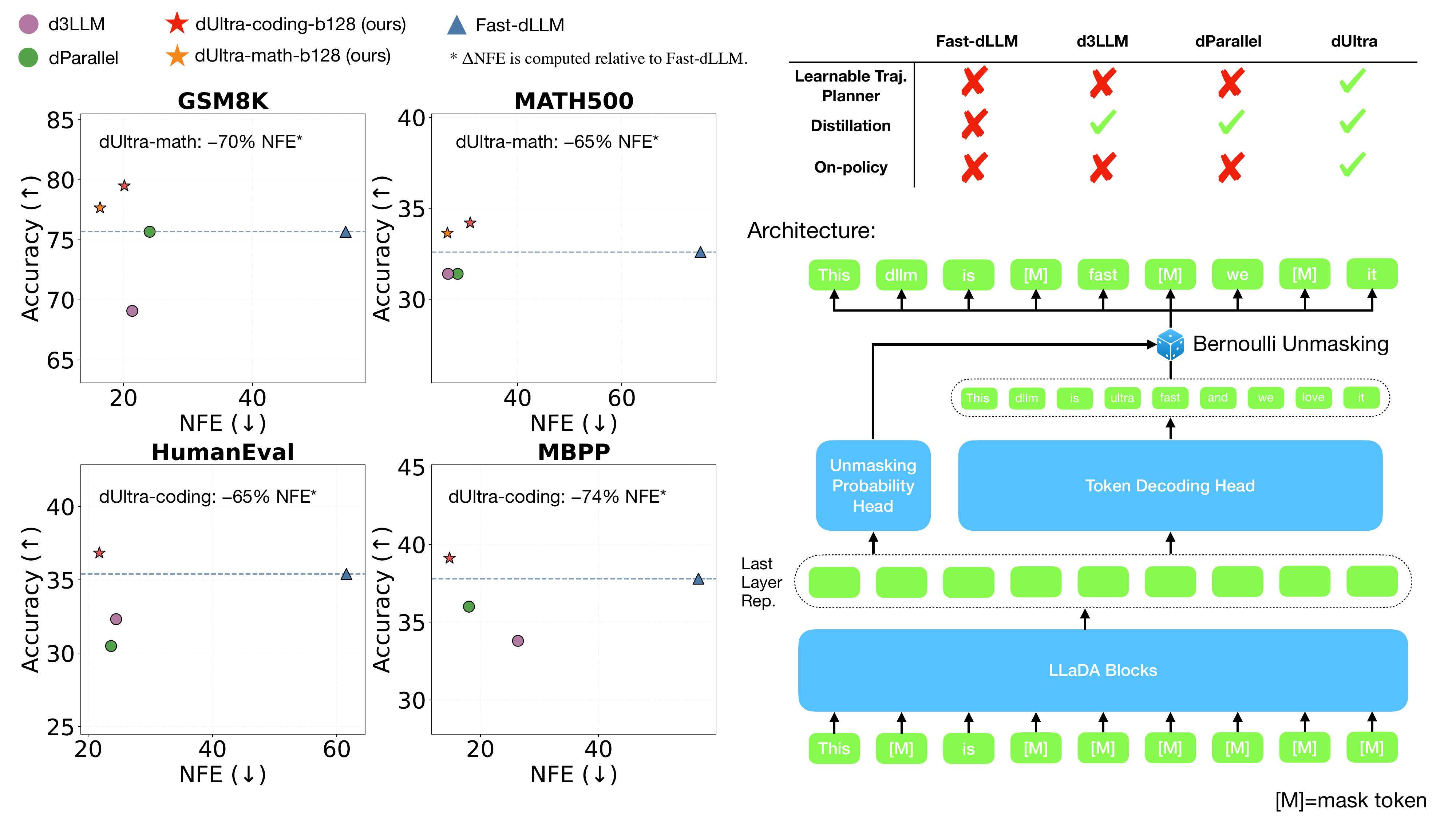}
   \caption{\textbf{dUltra overview and results.} Left: Accuracy versus number of function evaluations (NFE; denoising steps)
  on math and coding benchmarks for Fast-dLLM, dParallel, d3LLM, and dUltra (ours) variants. Here we use a block size of 128 and a generation length of 256. Right: dUltra architecture: a
  masked diffusion LM (LLaDA blocks) produces last-layer representations that feed (i) a token decoding head to output logits
  for masked positions and (ii) an unmasking planner head to output per-position unmasking probabilities; the Bernoulli sampled
  unmasking decisions update the masked sequence ([M] denotes the mask token).}
   \label{fig:main}
\end{figure}

Taking these motivations all together, in this work, we propose dUltra, a framework that \textit{learns} optimal unmasking strategies through on-policy RL combined with on-policy distillation reward. The primary result and architecture is summarized in \Cref{fig:main}. Our key contributions are:

\begin{itemize}
    \item We introduce an unmasking planner head that predicts per-token unmasking likelihood using independent Bernoulli distributions, enabling learnable sampling trajectory planning.
    \item We develop a GRPO-based on-policy distillation framework that jointly optimizes the diffusion model and unmasking order planner using both verifiable reward and distillation reward. We also propose advantage clipping to enable effective RL training of the unmasking planner head.
    \item We show state-of-the-art performance in terms of both token-per-forward (TPF) and number of function evaluations (NFE) on mathematical reasoning (GSM8K, MATH500) and code generation (HumanEval, MBPP) tasks.
\end{itemize} 

\section{Preliminary}
\subsection{Notation}
Let $\cX = \set{1, ..., m}$ be the support of the discrete distribution of a single token in the language sequence, where $m$ is the vocabulary size and the mask token is the $m$-th token. For any token $x\in \cX$, we denote $\bme_{x}\in \R^m$ as the one-hot probability vector corresponding to the degenerate distribution with $p(x) = 1$. Denote $\Delta^m = \set{\bs{\pi}\in \R^m \mid \sum_{i = 1}^m \pi_i = 1, \pi_i\geq 0}$ as the standard $m$-simplex, and $\Cat(\bpi)$ as the categorical distribution with the class probabilities $\bpi\in \Delta^m$.
\subsection{Masked Diffusion Language Model}
\subsubsection{Single token case}
Continuous-time MDLMs \citep{shi_simplified_2024,sahoo_simple_2024} can be formulated with a forward noising process and the corresponding reversal. For the forward noising process, we have for $t,s\in [0,1]$ and $t>s$,
\begin{equation}\label{eq:transition}
    \begin{split}
    q_{t|s}(x_t| x_s = m) &= \Cat(\bme_m), \\
    q_{t|s}(x_t|x_s = x_0) &= \Cat\left(\f{\alp_t}{\alp_s} \bme_{x_0} + \f{\alp_s - \alp_t}{\alp_s} \bme_m\right). 
    \end{split}
\end{equation}
where $\alp_t$ is the noise schedule function with $\alp_0 = 1$ and $\alp_1 = 0$. In particular,
\begin{equation}
    q_{t|0} (x_t | x_0) = \Cat(\alp_t \bme_{x_0} + (1-\alp_t) \bme_m),
\end{equation}
The time reversal for $s<t$ given $x_0$ is
\begin{equation} \label{eq:reverse}
    q_{s|t, 0} (x_s | x_t, x_0) = \f{q(x_t|x_s) q(x_s|x_0)}{q(x_t|x_0)} = \bcases{
    \Cat (\bme_{x_0}), & x_t=x_0 \\
    \Cat \left(\f{\alp_s -\alp_t}{1 - \alp_t}\bme_{x_0} + \f{1-\alp_s}{1-\alp_t}\bme_m\right) & x_t = m
    }
\end{equation}
Similar to DDPM \citep{ho_denoising_2020}, the parameterization of MDLM $\bs{\mu}_\tet$ aims to predict the clean sample $x_0$:
\begin{equation} \label{eq:param}
    p_{\bs{\tet}}(x_s | x_t) = q_{s|t, 0}(x_s | x_t, \bme_{x_0} \approx \bs{\mu}_\tet (x_t, t)).
\end{equation}

\subsubsection{Sequence case} \label{sec:seq}
For a token sequence $\bm{x}\in \cX^L$ of length $L$, MDLM factorizes the forward and reverse processes to sample each token independently. For the forward process, $q_{t|0} (\bm{x}_t | \bm{x}_0) = \prod_{l = 1}^L q_{t|0}(x_t^{(l)} | x_0^{(l)})$. For the reverse process, $q_{s|t, 0} (\bm{x}_s|\bm{x}_t, \bm{x}_0) = \prod_{l=1}^L q_{s|t,0}\left(x_s^{(l)}|x_t^{(l)}, x_0^{(l)} \approx \bs{\mu}_\tet^{(l)}(\bm{x}_t, t)\right)$. Note that here $\bs{\mu}_{\btet}$ is dependent on the entire sequence, so this factorization assumes conditional independence. 

\subsubsection{Likelihood bound}
The approximation in \Cref{eq:param} is more rigorously optimized by minimizing the negative ELBO (NELBO). Given a fixed number of discretization steps $T$, define $s(i) = (i-1)/T$ and $t(i) = i/T$. Then the NELBO can be decomposed as reconstruction loss, diffusion loss and prior loss, $\cL_{\textrm{recons}} + \cL_{\textrm{diffusion}} + \cL_{\textrm{prior}}$:
\begin{equation} \label{eq:nelbo}
\begin{split}   
\cL_{\textrm{NELBO}} = \mathbb{E}_{q} \Bigg[
    \underbrace{-\log p_{\theta}(x \mid x_{t(1)})}_{\mathcal{L}_{\text{recons}}}
    +
    \underbrace{\sum_{i=2}^{T} D_{\mathrm{KL}}\big[ q(x_{s(i)} \mid x_{t(i)}, x) 
    \,\|\, p_{\theta}(x_{s(i)} \mid x_{t(i)}) \big]}_{\mathcal{L}_{\text{diffusion}}}\\
    +
    \underbrace{D_{\mathrm{KL}}\big[ q(x_{t(T)} \mid x) 
    \,\|\, p_{\theta}(x_{t(T)}) \big]}_{\mathcal{L}_{\text{prior}}},
\Bigg]
\end{split}
\end{equation}
For MDLM with carry-over unmasking \citep{sahoo_simple_2024, shi_simplified_2024, nie2025largelanguagediffusionmodels, dream2025}, the NELBO loss can be simplified as 
\begin{equation} \label{eq:simple_loss}
    \cL_{\textrm{NELBO}} = -\int_0^1 \bbE_{q(x_t|x_0)}\left[\f{\alp_t'}{1 - \alp_t} \sum_{i = 0}^{\norm{\bm{x}}} \m{1}[\bm{x}_t^{(i)} = m] \cdot \log \bs{\mu_\tet}(\bm{x}_t;\bm{x}_t^{(i)} = \bm{x}_0^{(i)}) \right] dt,
\end{equation}
where the mean prediction network $\bs{\mu_\tet}(\bm{x}_t;\bm{x}_t^{(i)} = \bm{x}_0^{(i)})$ takes in the partially masked sequence $\bmx_t$ and predicts the likelihood of token $\bm{x}_t^{(i)}$ being the ground truth $\bm{x}_0^{(i)}$.
\subsubsection{Connection to any-order autoregressive modeling} \label{sec:ao-arm}
Any-order autoregressive models (AO-ARMs) \citep{uria2014deeptractabledensityestimator,hoogeboom2021argmaxflowsmultinomialdiffusion,shih2022traininginferenceanyorderautoregressive} evaluate the likelihood of a sequence $\bm{x}$ by marginalizing over all possible generation orders. More specifically, the expected negative log-likelihood minimized by AO-ARMs during training is
\begin{equation}
    \cL_{\theta}(\bm{x}) = -\bbE_{\xi \sim U(S_{\norm{\bm{x}}})}\left[\sum_{k=1}^{\norm{\bm{x}}} \log p_{\theta}(\bm{x}^{\xi(k)} | \bm{x}^{\xi(<k)})\right],
\end{equation}
where $U(S_{\norm{\bm{x}}})$ is the uniform distribution over all permutations over the initial mask tokens, and each permutation uniquely specifies a sampling order. Works have shown that this loss is equivalent to the NELBO loss of MDLMs with carry-over unmasking (\Cref{eq:simple_loss}) under mild conditions \citep{ou2025absorbingdiscretediffusionsecretly,kim_train_2025}. Therefore, MDLMs with carry-over unmasking can be interpreted as AO-ARMs that sample generation orders from a uniform distribution over all possible permutations.

\section{Methods}

\subsection{Path Planning with Unmasking Probability}

If we strictly follow the predetermined forward noising process in \Cref{eq:transition} and the corresponding time reversal in \Cref{eq:reverse} during denoising, a masked token has a \textit{fixed} probability of $\f{1-\alp_{s(i)}}{1-\alp_{t(i)}}$ remaining a mask token from time $t(i) = i/T$ to time $s(i) = (i-1)/T$ given the noise schedule function $\alp$. However, there is no inherent restriction on what the noise schedule function should be or what variable it should depend on under the framework of diffusion models. While maximizing ELBO requires sampling from the forward kernel $q(x_t|x_0)$ during training, at inference time we have flexibility in designing the reverse transition kernels $q_{s|t}$, allowing us to explore different unmasking strategies beyond the training distribution. One possibility is to make the schedule state-dependent by conditioning $\f{1-\alp_{s(i)}}{1-\alp_{t(i)}}$ on the current masked state $\bm{x}_t$, resulting in per-token schedules $\f{1-\bm{\alp}_{s(i)}^l(\bm{x}_t)}{1-\bm{\alp}_{t(i)}^l(\bm{x}_t)}$ for $l$-th token. In this way, the reverse denoising process is essentially a Markovian process with the transition kernel $p_{\bs{\tet}}(x_{s(i)} | x_{t(i)})$.

From the perspective of any-order autoregressive modeling, this is equivalent to learning a better distribution over sampling orders instead of the uniform distribution used in the standard negative log-likelihood. As motivated in \Cref{fig:case_study}, we would like to refine this distribution to concentrate the probability mass over sampling orders that maximize the average number of conditionally independent tokens to unmask per step. Therefore, we want to learn a better unmasking order planner that can decide which tokens to unmask at each denoising step.

We thereby consider having an unmasking order planner to decide which mask tokens to unmask and how likely we will unmask them. To decide which tokens to unmask at each step, we model per-token unmasking decisions as independent Bernoulli distributions, mirroring the conditional independence assumption in MDLM's token prediction factorization (\Cref{sec:seq}). Modeling the full joint distribution over all possible position subsets would be intractable due to the exponential number of possibilities ($2^n$ subsets for n masked tokens). While we sample unmasking decisions independently for computational feasibility, the planner can still capture dependencies between positions through self-attention when computing the unmasking probabilities. We introduce an unmasking planner head $\phi$ that takes the last-layer hidden representations of the base MDLM and outputs per-token unmasking probabilities:
\begin{equation}
p_{\text{unmask}}(\bm{x}) \sim \Ber(\sigmoid(\phi(h_{\bm{x}}, \bm{x})))
\end{equation}
where $h_{\bm{x}}$ is the last-layer hidden representation of the sequence of tokens $\bm{x}$.

The unmasking planner head $\phi$ is a lightweight module consisting of a single transformer block (matching the base MDLM architecture), time embedding layers for timestep conditioning, adaptive layer normalization, and a final linear projection to scalar logits. This design keeps the computational overhead minimal while allowing the planner to capture dependencies between masked positions and incorporate temporal information when deciding which tokens to unmask. The planner processes all masked token representations in parallel and outputs conditionally independent unmasking probabilities for each token. See \Cref{app:arch} for complete architectural details.

\subsection{Exact Likelihood Calculation} \label{sec:likelihood}
One of the advantages of introducing an explicit unmasking planner head is that we can tractably compute the likelihood of each rollout during training, which is essential for policy gradient methods. Recall from the AO-ARM perspective (\Cref{sec:ao-arm}) that MDLMs define a Markovian generative process: at each step, we unmask a subset of tokens conditioned on all previously unmasked tokens. This Markovian structure makes the likelihood calculation straightforward---we simply multiply the transition probabilities at each step. Next we compute the transition probability at each denoising step.

We denote the prompt as $\m{q}$ and the sequence completion after denoising step $k$ as $\m{o}^k$, and $\m{o}^0$ is the completely masked sequence. At step $k$, we select positions $\textbf{Pos}^{(k)}\subseteq\cX_{\text{mask}}^{(k)}$ to unmask from the set of currently masked tokens $\cX_{\text{mask}}^{(k)}$. Our model consists of two key components: (1) $\pi_{\text{token}}$, the token probability distribution for masked positions, which we get from the logits of the model output, and (2) $\pi_{\text{select}}$, the unmasking order planner that unmasks each token with probability $p_{\text{unmask}}$ at each step. With these components, we model the parallel decoding procedure as an ancestral sampling process. We first sample the positions to unmask from $\pi_{\text{select}}$, and then sample the tokens for the selected positions from $\pi_{\text{token}}$. Therefore, the likelihood of generating $\bm{o}^k$ from $\bm{o}^{k-1}$ given prompt $\bm{q}$ is:
\begin{equation} \label{eq:likelihood}
p\big(\bm{o}^k \mid \bm{q}, \bm{o}^{k-1}\big)
= \pi_{\text{select}}\big(\textbf{Pos}^{(k)} \mid \bm{q}, \bm{o}^{k-1}\big)
\cdot \pi_{\text{token}}\big(\bm{o}^k \mid \bm{q}, \bm{o}^{k-1}, \textbf{Pos}^{(k)}\big)
\end{equation}

The token selection probability factorizes as:
\[
\pi_{\text{token}}\big(\bm{o}^k \mid \bm{q}, \bm{o}^{k-1}, \textbf{Pos}^{(k)}\big)
= \prod_{j=1}^{\#\text{unmask}}
\pi_\theta\big(\bm{o}_{\textbf{Pos}^{(k)}_j}^k \mid \bm{q}, \bm{o}^{k-1}, \textbf{Pos}^{(k)}\big)
\]

The position selection probability follows the product of independent Bernoulli distributions:
\begin{equation}
    \pi_{\text{select}}\big(\textbf{Pos}^{(k)} \mid \bm{q}, \bm{o}^{k-1} \big)
    = \prod_{x \in \textbf{Pos}^{(k)}} p_{\text{unmask}}(x)
    \cdot \prod_{x \in \cX_{\text{mask}}^{(k)} \setminus \textbf{Pos}^{(k)} } \big(1 - p_{\text{unmask}}(x)\big)
\end{equation}
In practice, one can additionally incorporate a block size hyperparameter $B$ to restrict the set of candidate positions: instead of selecting from all masked tokens $\cX_{\text{mask}}^{(k)}$, we constrain $\textbf{Pos}^{(k)}$ to be a subset of the leftmost $B$ masked tokens. The block size interpolates between completely learnable unmasking order and autoregressive-like order: larger $B$ allows the planner to select from positions throughout the sequence, while smaller $B$ restricts candidates to the leftmost masked tokens, enforcing a more left-to-right progression similar to autoregressive decoding, though the planner retains flexibility in ordering tokens within each block.

Hence, the overall likelihood of a complete rollout with $K$ denoising steps is:
\begin{equation} \label{eq:p_student}
p\big(\bm{o}^{K} \mid \bm{q}\big)
= \prod_{k=1}^K p\big(\bm{o}^k \mid \bm{q}, \bm{o}^{k-1}\big)
= \prod_{k=1}^K \pi_{\text{select}}\big(\textbf{Pos}^{(k)} \mid \bm{q}, \bm{o}^{k-1}\big)
\cdot \pi_{\text{token}}\big(\bm{o}^k \mid \bm{q}, \bm{o}^{k-1}, \textbf{Pos}^{(k)}\big).
\end{equation}
Equivalently, we can view each token's unmasking trajectory as following a non-homogeneous geometric distribution\footnote{Unlike the standard geometric distribution with constant success probability, the non-homogeneous variant allows the probability to vary across steps.}: at each denoising step, the planner assigns a state-dependent probability of unmasking, and the token remains masked until it is eventually sampled to be unmasked.
\subsection{Initializing on-policy scheduler with heuristic methods}
We find that if we initialize the unmasking planner head randomly, the expected number of tokens unmasked at each denoising step $n = \sum_i p_{\text{unmask}}(x_i)$ can be excessively large, leading to degenerate performance at the onset of training, preventing the RL algorithm from effectively exploring sampling trajectories that have more sampling steps and better performance. Therefore, we propose to initialize the planner before applying RL so that the RL algorithm uses the heuristic logit-based planner as a baseline to optimize the sampling order. 

More specifically, we finetune the unmasking planner head using supervised learning to mimic a confidence-based sampling strategy. We train the planner to predict token selection decisions made by the confidence-aware parallel decoding method proposed in Fast-dLLM \citep{wu2025fastdllmtrainingfreeaccelerationdiffusion}. This method selects highly confident tokens to unmask in each denoising step. This avoids the pathological case described in \Cref{fig:bimodal}, and provides a strong baseline for our learnable planner.

We formulate the planner finetuning as a binary classification problem: given a partially masked input, the planner must predict which tokens should be unmasked at the current step according to the Fast-dLLM confidence criterion. For a set of masked positions $\mathcal{I}_{\text{block}}$, we optimize the unmasking planner head $\phi$ using binary cross-entropy loss with position weighting to handle class imbalance:
\begin{equation}
    \mathcal{L}_{\text{planner}} = \frac{1}{|\mathcal{I}_{\text{block}}|} \sum_{i \in \mathcal{I}_{\text{block}}} \text{BCE}\big(\phi_i(\bm{h}), y_i; w_{\text{pos}} = \frac{\#\text{negatives}}{\#\text{positives}}\big)
\end{equation}
where $\phi_i(\bm{h})$ is the predicted unmasking probability for $i$-th masked token, $y_i \in \{0,1\}$ indicates whether token $i$ should be unmasked according to Fast-dLLM, and $w_{\text{pos}}$ weights the loss for tokens that should be unmasked to compensate for the class imbalance between tokens to unmask and tokens to keep masked.
The finetuned planner can then be further optimized using GRPO to discover a more optimized sampling order beyond the fixed confidence-based heuristic.

\subsection{On-policy distillation reward}

As motivated by \Cref{fig:bimodal}, to further accelerate sampling with parallel decoding beyond sampling conditionally independent tokens, the model must learn to sample from a single high-quality mode among multiple plausible modes to avoid generating incoherent text, i.e. the model should exhibit mode-seeking behavior. To encourage this behavior, we introduce an on-policy distillation reward that encourages the student diffusion LLM to maximize the sequence-level log-likelihood of on-policy samples evaluated by an autoregressive teacher model.

We use domain-specific teacher models to provide high-quality supervision signals: Qwen2.5-Math-7B-Instruct \citep{qwen2.5math} for mathematical reasoning and Qwen2.5-Coder-7B-Instruct \citep{qwen2.5coder} for code generation. Given a prompt $\bm{q}$ and a completion $\bm{c}$ generated by the student model during rollout, we compute the on-policy distillation reward as:
\begin{equation}
r_{\text{distill}}(\bm{q}, \bm{c}) = \frac{1}{|\bm{c}|} \left(\left[\sum_{i=1}^{|\bm{c}|} \log p_{\text{teacher}}(c_i \mid \bm{q}, \bm{c}_{<i})\right] - \beta \log p_{\text{student}}(\bm{c} \mid \bm{q})\right)
\end{equation}
where $|\bm{c}|$ is the length of the completion in tokens, $p_{\text{student}}(\bm{c} \mid \bm{q})$ is given by \Cref{eq:p_student}, and $\beta$ controls the weight of the entropy regularization term. When $\beta =1$, maximizing $\bbE_{\bm{c}\sim p_{\text{student}}}[r_{\text{distill}}]$ corresponds to minimizing a scaled reverse KL divergence $\frac{1}{|\bm{c}|} D_{\text{KL}}(p_\text{student} \| p_\text{teacher})$. Reverse KL divergence is known to encourage mode-seeking behavior \citep{minka2005divergence}: when the student places probability mass where the teacher model has low probability, the term $\log(p_{\text{student}}/p_{\text{teacher}})$ becomes large and heavily penalizes the divergence. We set $\beta = 0$ in our experiments, simplifying the reward to average teacher log-likelihood, while still encouraging mode-seeking behavior: the student learns to generate completions that the teacher assigns high probability, focusing on high-quality modes verified by the teacher model rather than exploring low-probability regions that may lead to incoherent text. Empirically, we did not find significant differences in performance when varying $\beta$.

The on-policy distillation reward is combined with verifiable rewards and efficiency reward (i.e., negative number of denoising steps) via a weighted sum:
\begin{equation}
r_{\text{total}} = \lambda_{\text{task}} \cdot r_{\text{task}} + \lambda_{\text{distill}} \cdot r_{\text{distill}} + \lambda_{\text{step}} \cdot r_{\text{step}}
\end{equation}
where the weights $\lambda_{\text{task}}, \lambda_{\text{distill}}, \lambda_{\text{step}}$ control the relative importance of each component. This multi-objective reward encourages the model to generate correct, coherent, and efficiently-sampled outputs. See \Cref{app:reward} for detailed reward configuration for each task.

\subsection{GRPO Training Framework}
We train the diffusion LLM and unmasking planner head using Group Relative Policy Optimization (GRPO), a policy-gradient method that operates on groups of trajectories per prompt and uses group-relative advantages as baselines. We use on-policy updates (\texttt{num\_iterations}$=1$) as off-policy updates provide no computational benefit: unlike autoregressive LLMs that evaluate likelihoods in a single forward pass, each likelihood evaluation using \Cref{eq:likelihood} requires a complete denoising process. For each group, we sample a single prompt $\bm{q}$ and generate $G$ trajectories with denoised sequence at step $k$ of $g$-th trajectory denoted as $\bm{o}^{g,k}$. The group-relative advantages of these trajectories are calculated as $A^g = r^g - \text{mean}(r^{1:G})$. Note that we do not normalize the advantage to avoid introducing any bias into the policy gradient estimation \citep{liu2025understanding}. Then the GRPO objective reduces to the REINFORCE-style loss
\begin{equation}
    \cL_{\text{GRPO}}(\theta) = - \frac{1}{\norm{\bm{o}}\cdot G} \sum_{g=1}^G A^g \sum_{k=1}^{K^{g}} \log p_{\theta}\big(\bm{o}^{g,k} \mid \bm{q}, \bm{o}^{g,k-1}\big),
\end{equation}
where $p_{\theta}(\bm{o}^{g,k} \mid \bm{q}, \bm{o}^{g,k-1})$ is the rollout likelihood defined in \Cref{eq:likelihood}, and $K^g$ is the total number of denoising steps for $g$-th trajectory. In practice, we use the ratio formulation $\frac{p_{\theta}}{\text{sg}(p_{\theta})}$ instead of $\log p_{\theta}$ for numerical stability, where $\text{sg}(\cdot)$ denotes stop-gradient. The two formulations have identical gradients. 

\textbf{Advantage clipping.} We find that using vanilla GRPO leads to the degenerate solution where all unmasking probabilities collapse to 0, even with the accurate likelihood computed in \Cref{sec:likelihood}. While this gives high distillation reward due to extremely slow sampling, it defeats the purpose of accelerating sampling. We hypothesize that this occurs because trajectories with low advantage may be overly conservative—unmasking fewer tokens per step than the model could have safely unmasked in parallel. For example, if the model could have unmasked 5 tokens but the planner only unmasked 3, the trajectory receives low advantage due to inefficiency. However, when GRPO penalizes this trajectory, it uniformly reduces the likelihood of all unmasking decisions, including the 3 tokens that were correctly unmasked. This creates noisy gradients that penalize correct unmasking decisions along with the overly conservative behavior, and we observe the unmasking planner head quickly collapses to never unmasking any tokens.

To address this, we introduce advantage clipping that selectively focuses learning on successful trajectories. After computing advantages $A^g$ for each rollout $g$ in a group, we apply a lower threshold:
\begin{equation}
A^g \gets \begin{cases}
0 & \text{if } A^g < C \\
A^g & \text{otherwise}
\end{cases}
\end{equation}
where $C$ is the clipping threshold. This effectively ignores trajectories that performed significantly worse than average, allowing the planner to learn primarily from successful unmasking strategies. Another intuition is that the space of all unmasking orders is exponentially large compared to valid unmasking orders, making it difficult to learn from failed attempts, and it is probably more effective to explore for successful unmasking orders and exploit them. Empirically, we find that setting $C$ to $0$ stabilizes training.

In summary, the complete training algorithm can be found in Appendix \ref{app:alg}.

\section{Results}

We evaluate dUltra on mathematical reasoning and code generation benchmarks to demonstrate its ability to achieve competitive performance with significantly reduced computational cost. All of our experiments use LLaDA-8B-Instruct as the base model. We compare against \textit{state-of-the-art} confidence-based heuristics, such as Fast-dLLM \citep{wu2025fastdllmtrainingfreeaccelerationdiffusion}, and \textit{state-of-the-art} off-policy distillation-based methods, such as d3LLM \citep{d3llm} and dParallel \citep{chen2025dparallellearnableparalleldecoding}. We report both task performance and the average number of function evaluations (NFE), i.e., the number of denoising steps during generation.
\subsection{Training Dynamics}

\begin{figure}[h]
\centering
\includegraphics[width=\textwidth]{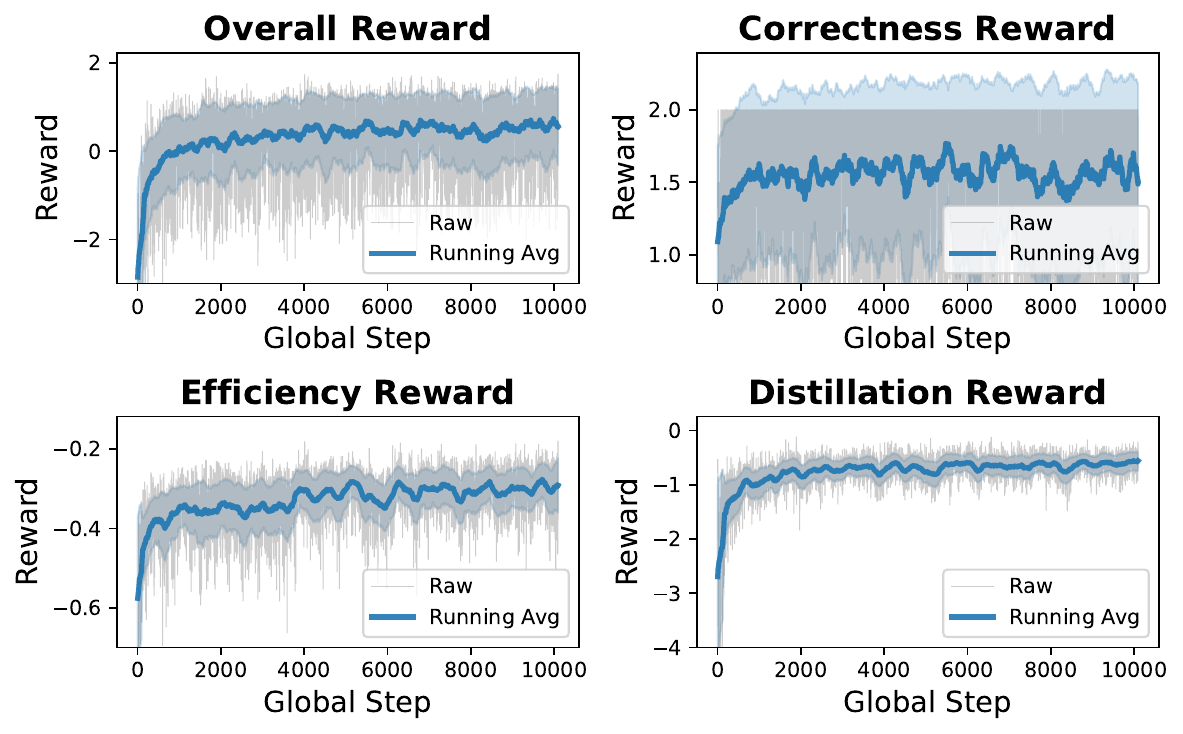}
\caption{Training dynamics on GSM8K.}
\label{fig:training_curves}
\end{figure}

\Cref{fig:training_curves} shows the evolution of reward metrics during dUltra training on GSM8K. All three rewards, including correctness reward, efficiency reward, and distillation reward, increase stably throughout training. The correctness reward is measured by the correctness of generated solutions, and receives 2.0 if the answer is correct and 0.0 otherwise. The efficiency reward is defined as the negative number of denoising steps (NFE) times a constant. The distillation reward is the log likelihood of rollouts with the teacher model. We observe that the efficiency increases rapidly with training onset, enabling efficient rollouts throughout training. Overall, these training dynamics validate the effectiveness of our GRPO framework and multi-objective reward design in guiding the diffusion LLM towards generating high-quality and efficient outputs.

\begin{figure}
\centering
\includegraphics[width=0.9\textwidth]{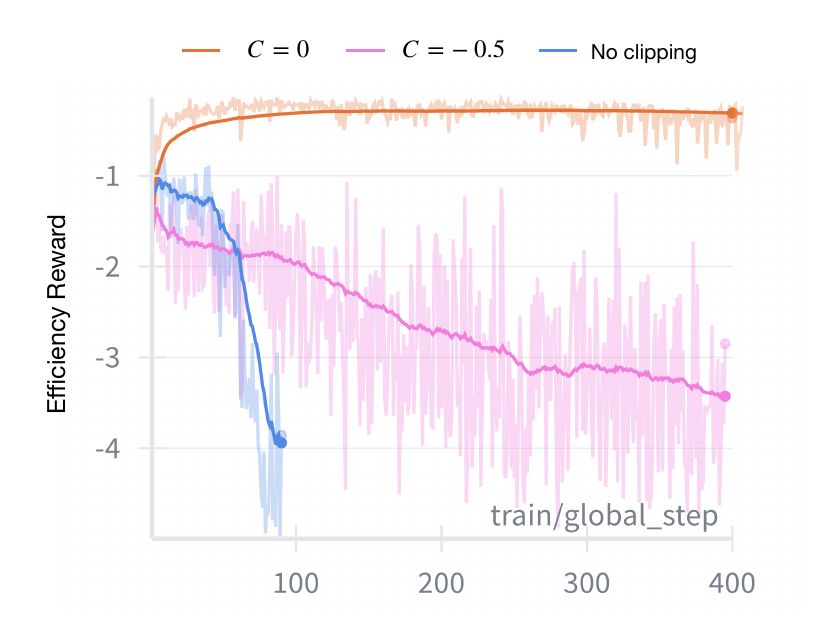}
\caption{Effect of advantage clipping value $C$ on training stability. Without advantage clipping, the average NFE quickly diverges as the planner collapses to never unmasking any tokens. With advantage clipping, training remains stable and the model learns to reduce NFE while maintaining task performance.}
\label{fig:adv_clip}
\end{figure}

\textbf{Effect of Advantage Clipping on Efficiency Reward.}
 The role of advantage clipping in training stability is illustrated in \Cref{fig:adv_clip}. Without advantage clipping ($C = -\infty$), the training quickly becomes unstable: the average NFE diverges to extreme values as the learned planner collapses to a degenerate policy that never unmasks tokens. This occurs because negative advantages from poor trajectories can dominate the gradient signal, causing the planner to become overly conservative. With appropriate advantage clipping ($C = 0.0$), training remains stable and the model learns a balanced unmasking strategy. The clipping mechanism ensures that only rollouts with above-average returns contribute to policy updates, preventing the model from over-fitting to failure modes. This ablation validates our design choice and highlights the importance of careful advantage normalization in GRPO training for diffusion models.

\subsection{Main Results}
\subsubsection{Results with Small Block Size}

We first evaluate dUltra with block size $B=32$, which provides a balance between sampling efficiency and unmasking selectivity. All tasks are evaluated in a zero-shot setting. Since sampling using the unmasking order planner is stochastic, we average over 5 trials for the result of dUltra models with planner. \Cref{tab:b32_results} presents our results on mathematical reasoning (GSM8K, MATH500) and code generation (HumanEval, MBPP) tasks, comparing against multiple accelerated MDLMs based on distillation, including dParallel and d3LLM. We report both task performance and the average number of function evaluations (NFE), i.e., the number of denoising steps during generation.

\begin{table}[h]
\centering
\scriptsize
\begin{tabular}{lcccccccc}
\toprule
& \multicolumn{4}{c}{\textbf{Math}} & \multicolumn{4}{c}{\textbf{Coding}} \\
\cmidrule(lr){2-5} \cmidrule(lr){6-9}
& \multicolumn{2}{c}{\textbf{GSM8K}} & \multicolumn{2}{c}{\textbf{MATH500}} & \multicolumn{2}{c}{\textbf{HumanEval}} & \multicolumn{2}{c}{\textbf{MBPP}} \\
\cmidrule(lr){2-3} \cmidrule(lr){4-5} \cmidrule(lr){6-7} \cmidrule(lr){8-9}
\textbf{Method} & Acc$\uparrow$ & NFE$\downarrow$ (T/F$\uparrow$) & Acc$\uparrow$ & NFE$\downarrow$ (T/F$\uparrow$) & Acc$\uparrow$ & NFE$\downarrow$ (T/F$\uparrow$) & Acc$\uparrow$ & NFE$\downarrow$ (T/F$\uparrow$) \\
\midrule
Fast-dLLM & 77.56 & 57.26 (4.33) & \textbf{37.20} & 77.12 (3.22) & 33.54 & 62.65 (3.94) & \textbf{37.80} & 50.04 (4.16) \\
dParallel & 77.10 & 28.47 (7.17) & 31.80 & 35.21 (5.32) & 31.71 & 30.68 (5.87) & 36.20 & 22.77 (6.30) \\
d3LLM & 72.50 & 25.79 (8.39) & 32.60 & 32.04 (6.81) & \textbf{37.20} & 33.41 (6.86) & 33.20 & 32.26 (6.90) \\
\textbf{dUltra-math (ours)} & \underline{81.29} & 21.21 (9.717) & 35.52 & 33.20 (7.03) & - & - & - & - \\
\ \ \ \ \ \ \ w/o planner & 75.44 & \underline{15.31} (13.39) & 34.00 & \underline{24.67} (9.45) & -&-&-&- \\
\textbf{dUltra-coding (ours)} & \textbf{81.52} & 23.54 (8.40) & \underline{35.64} & 33.39 (6.72) & \underline{35.85} & \underline{24.29} (7.23) & \underline{37.04} & \underline{19.78} (7.29) \\
\ \ \ \ \ \ \ w/o planner & 75.89& \textbf{15.27} (12.84)&  32.60&  \textbf{22.49} (9.89)& 23.78&\textbf{16.18} (10.19)&20.40&\textbf{12.18} (11.28) \\
\bottomrule
\end{tabular}
\vspace{1em}
\caption{Main results comparing dUltra with baseline methods across mathematical reasoning and code generation tasks. The generation length is 256 and the block size is 32 for all results in this table. We report accuracy, NFE (number of function evaluations), and T/F (tokens per forward pass) for each task. \textbf{Bold} indicates the best result and \underline indicates the second-best result for each column. The ``w/o planner'' rows show results using Fast-dLLM sampling with the base model's logits (without the learned unmasking planner head), serving as an ablation baseline. dUltra-math is trained on GSM8K training dataset while dUltra-coding uses dUltra-math as the base checkpoint and is trained on the APPS dataset.}
\label{tab:b32_results}
\end{table}

dUltra achieves strong performance across all benchmarks while maintaining low NFE. Notably, dUltra outperforms both distillation baselines (d3LLM and dParallel) on GSM8K and MATH500 in accuracy, while using fewer denoising steps on GSM8K and comparable denoising steps on MATH500, consistent with our mode filtering hypothesis: on-policy RL can discover better modes than methods limited to off-policy teacher trajectories. Comparing the full GRPO-trained system using its learned unmasking order planner against confidence-based sampling on the same base model (w/o planner), we observe consistent improvements: +5.9 pp on GSM8K and +12--17 pp on code generation tasks. This demonstrates the value of jointly training the base model and unmasking order planner for task-specific unmasking strategies.

\subsubsection{Results with Large Block Size}

To explore the impact of larger block sizes on sampling efficiency, we evaluate dUltra with $B=128$. Larger block sizes give the planner more freedom for longer-horizon planning by allowing it to select from a broader set of candidate positions at each denoising step, potentially enabling faster inference through increased parallelism. \Cref{tab:b128_results} presents the results with block size 128 with all other settings the same as results with block size 32.

\begin{table}[h]
\centering
\scriptsize
\begin{tabular}{lcccccccc}
\toprule
& \multicolumn{4}{c}{\textbf{Math}} & \multicolumn{4}{c}{\textbf{Coding}} \\
\cmidrule(lr){2-5} \cmidrule(lr){6-9}
& \multicolumn{2}{c}{\textbf{GSM8K}} & \multicolumn{2}{c}{\textbf{MATH500}} & \multicolumn{2}{c}{\textbf{HumanEval}} & \multicolumn{2}{c}{\textbf{MBPP}} \\
\cmidrule(lr){2-3} \cmidrule(lr){4-5} \cmidrule(lr){6-7} \cmidrule(lr){8-9}
\textbf{Method} & Acc$\uparrow$ & NFE$\downarrow$ (T/F$\uparrow$) & Acc$\uparrow$ & NFE$\downarrow$ (T/F$\uparrow$) & Acc$\uparrow$ & NFE$\downarrow$ (T/F$\uparrow$) & Acc$\uparrow$ & NFE$\downarrow$ (T/F$\uparrow$) \\
\midrule
Fast-dLLM & 75.66 & 54.43 (4.56) & 32.60 & 75.72 (3.30) & \underline{35.40} & 61.54 (4.08) & \underline{37.80} & 56.93 (4.07) \\
dParallel & 75.66 & 24.07 (8.23) & 31.40 & 28.49 (5.45) & 30.49 & 23.68 (6.26) & 36.00 & 18.07 (7.63) \\
d3LLM & 69.07 & 21.37 (9.59) & 31.40 & 26.65 (7.03) & 32.32 & 24.49 (7.34) & 33.80 & 26.37 (7.66) \\
\textbf{dUltra-math-b128 (ours)} & \underline{77.65} & 16.38 (13.96) & \underline{33.64} & 26.53 (9.25) & - & - & - & - \\
\ \ \ \ \ \ \ w/o planner & 69.67& \textbf{10.46} (22.16)& 32.80  & \underline{19.36} (12.80)  & -&-&-&- \\
\textbf{dUltra-coding-b128 (ours)} & \textbf{79.48} & 20.15 (9.99) & \textbf{34.20} & 30.90 (7.50) & \textbf{36.83} & \underline{21.79} (9.43) & \textbf{39.12} & \underline{14.78} (9.97) \\
\ \ \ \ \ \ \ w/o planner & 71.57& \underline{10.90} (18.44)&  30.60& \textbf{18.46} (12.50) & 21.95&\textbf{13.18} (15.00)&21.60&\textbf{8.27} (17.65) \\
\bottomrule
\end{tabular}
\vspace{1em}
\caption{Main results comparing dUltra with baseline methods across mathematical reasoning and code generation tasks. The generation length is 256 and the block size is 128 for all results in this table. We report accuracy, NFE (number of function evaluations), and T/F (tokens per forward pass) for each task. \textbf{Bold} indicates the best result and \underline indicates the second-best result for each column. The ``w/o planner'' rows show results using Fast-dLLM sampling with the base model's logits (without the learned unmasking planner head), serving as an ablation baseline. dUltra-math-b128 is trained on GSM8K training dataset with rollouts using block size of 128, and uses dUltra-math as the base checkpoint. dUltra-coding-b128 is trained on the APPS dataset with rollouts using block size of 128, and uses dUltra-coding as the base checkpoint.}
\label{tab:b128_results}
\end{table}

Larger block sizes enable faster sampling with minimal accuracy loss. dUltra-math-b128 achieves 16.38 NFE on GSM8K (23\% reduction vs. $B=32$) with only 3.6 pp accuracy drop. The full system using the learned unmasking order planner shows consistent improvements over confidence-based sampling on the same base model ($\sim$8 pp on GSM8K and $\sim$15--18 pp on code tasks). This demonstrates that joint training enables effective adaptation to larger block sizes.

\subsection{Speed-Quality Trade-off}
\begin{figure}[h]
\centering
\includegraphics[width=\textwidth]{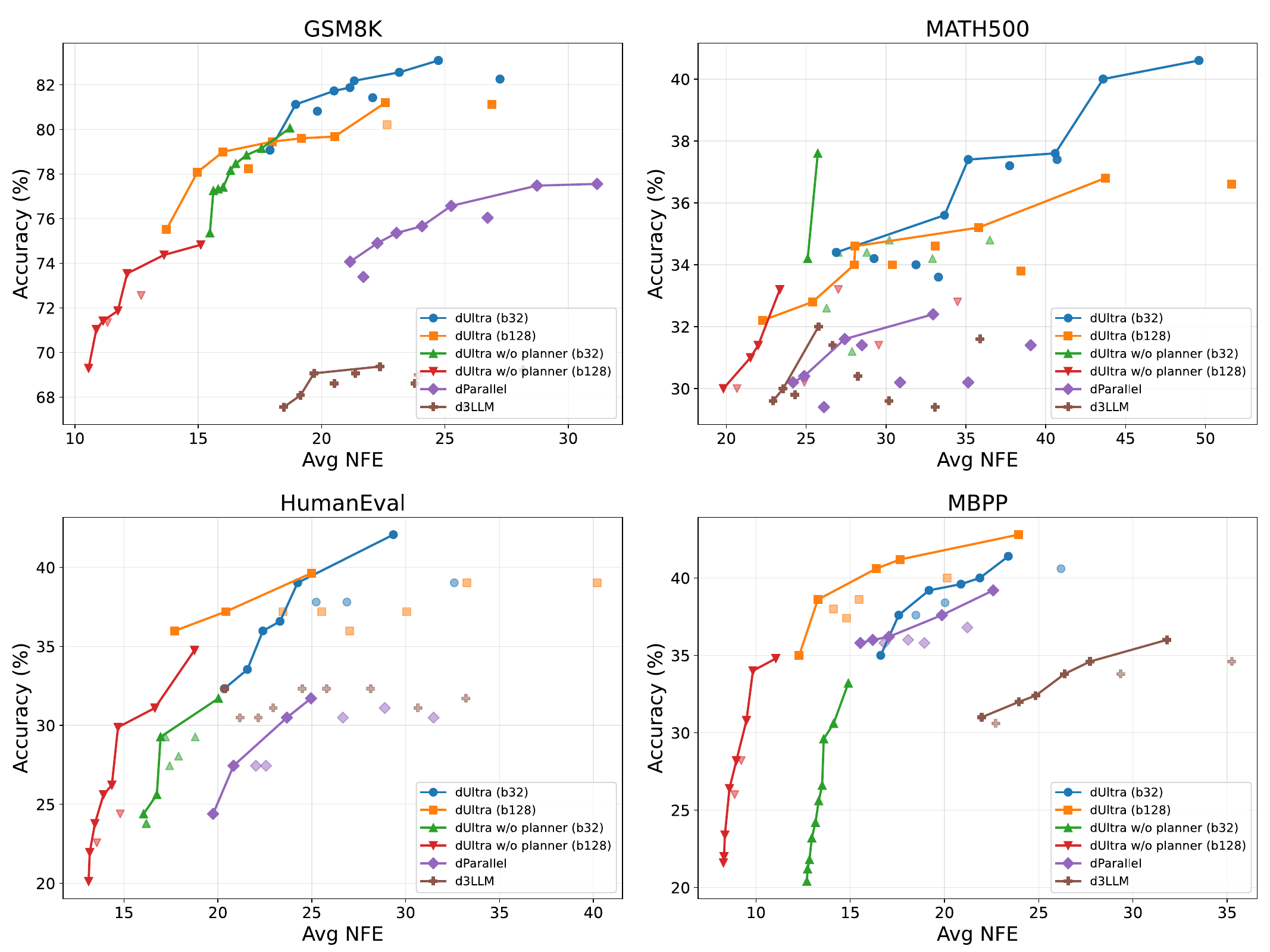}
\caption{Accuracy-efficiency Pareto frontiers across math reasoning and code generation benchmarks. We
   compare variants of dUltra against baseline methods on GSM8K
  (top-left), MATH500 (top-right), HumanEval (bottom-left), and MBPP (bottom-right). The x-axis shows
  average NFE (lower is faster), while the y-axis shows task accuracy. Each point represents a
  different threshold for dUltra models with planners and a different factor for baseline models without planners (including dUltra models without planners). dUltra models with planner consistently achieves superior accuracy-efficiency trade-offs compared to distillation-based methods, demonstrating the effectiveness of our learned unmasking strategies via on-policy distillation.}
\label{fig:pareto_threshold}
\end{figure}

During inference time, to eliminate randomness and dynamically adjust the sampling speed, we use a deterministic threshold to only unmask tokens with unmasking probability above that threshold. This allows us to strike a balance between sampling efficiency (NFE) and task performance (accuracy). We use Fast-dLLM's factor-based parallel decoding strategy \citep{wu2025fastdllmtrainingfreeaccelerationdiffusion} for models without a planner. We sweep across the same set of thresholds (for dUltra models) and factors (for models w/o a planner) and show the Pareto frontier in \Cref{fig:pareto_threshold}. We also show that continual training of dUltra-math on coding tasks does not alter the Pareto curve significantly (see \Cref{fig:pareto_mathcoding}). Furthermore, we explore another trade-off strategy with a multiplicative factor applied to the unmasking probability in Appendix \ref{app:trade-off}.

\subsection{Analysis of Learned Unmasking Strategies}

\begin{figure}[h]
\centering
\includegraphics[width=\textwidth]{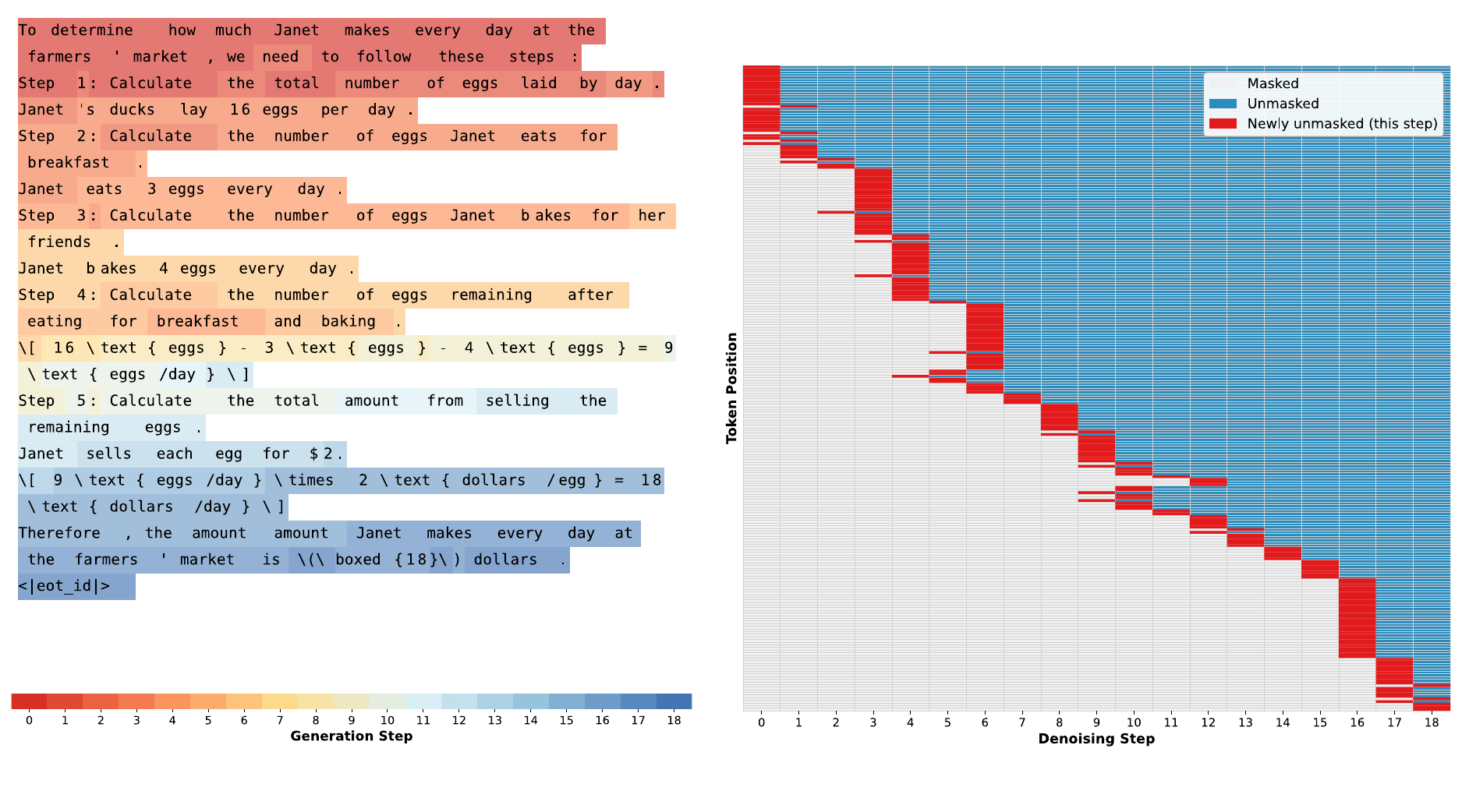}
\caption{Visualization of learned unmasking patterns on GSM8K. Red bars show newly unmasked tokens at each denoising step. The pattern exhibits a predominantly autoregressive (left-to-right) trend with minor deviations. An interesting observation is that sometimes, "Step" and ":" are generated first without the numbering in between, and the numbering is generated in the next step. }
\label{fig:math_order}
\end{figure}

\begin{figure}[h]
\centering
\includegraphics[width=\textwidth]{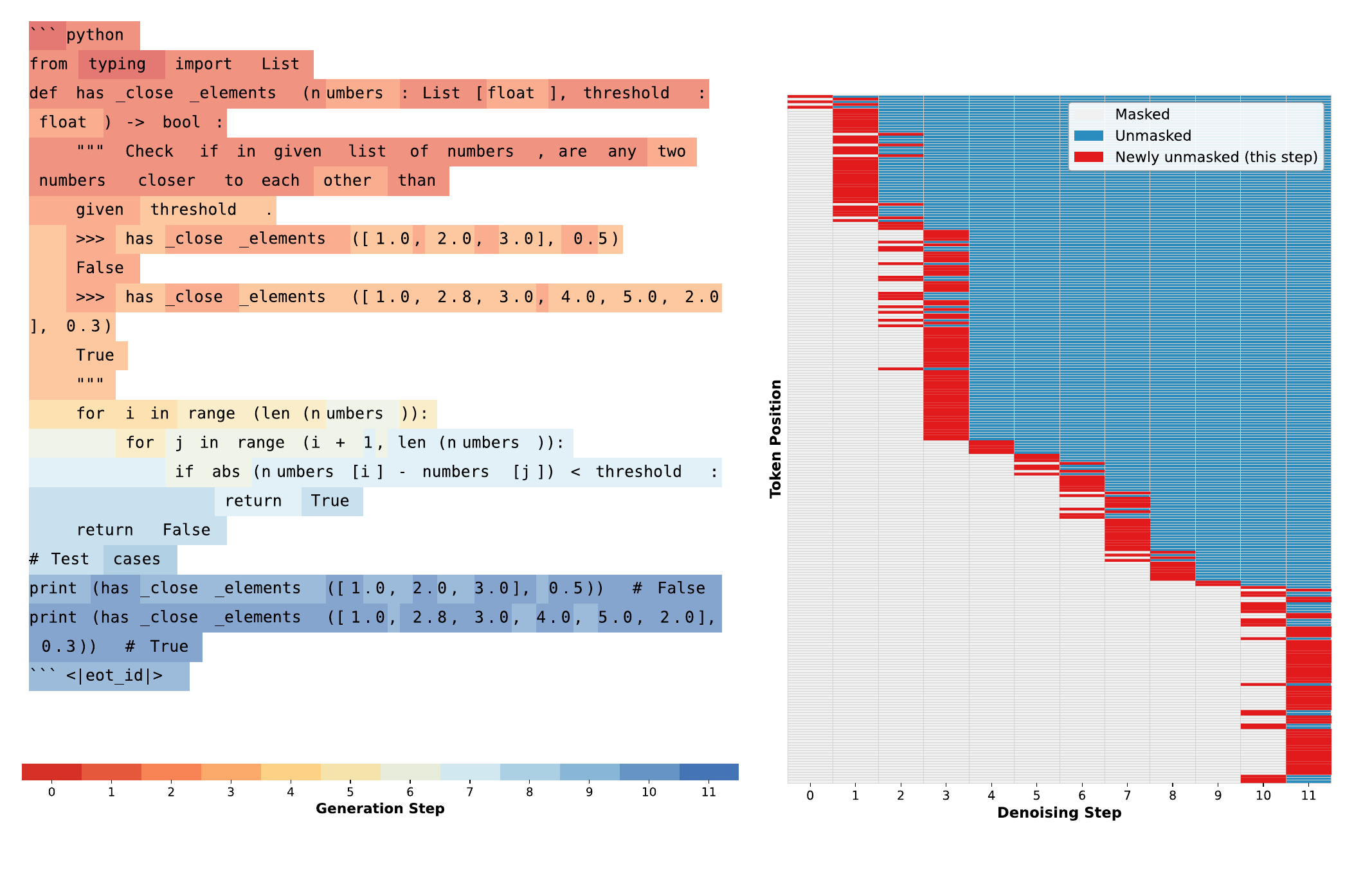}
\caption{Visualization of learned unmasking patterns on HumanEval. While the overall trend remains autoregressive, code generation exhibits more non-autoregressive behavior with visible jumps to unmask function signatures, control flow keywords, and structural elements earlier than strict left-to-right order.}
\label{fig:coding_order}
\end{figure}

To understand what unmasking order patterns dUltra learns, we visualize the unmasking behavior across denoising steps on both math and coding tasks (\Cref{fig:math_order} and \Cref{fig:coding_order}). Each heatmap shows when tokens are unmasked during the denoising process, with red bars indicating newly unmasked tokens at each step. The vertical axis represents token positions, while the horizontal axis shows denoising steps. For both visualization, we use a block size of 128 to allow more flexible unmasking orders.

The visualizations reveal that the major trend across both tasks is largely autoregressive: tokens are predominantly unmasked in left-to-right order, as evidenced by the diagonal progression from top-left to bottom-right in both heatmaps. This suggests that despite the theoretical flexibility of masked diffusion models, the autoregressive order often serves as a good enough prior. This also explains the success of AR models and speculative decoding in practice.

Nevertheless, we observe that code generation exhibits more non-autoregressive behavior compared to mathematical reasoning. In \Cref{fig:coding_order}, we observe visible jumps where the model unmasks function signatures (\texttt{def has\_close\_elements}), control flow keywords (\texttt{for}, \texttt{if}), and return statements earlier than strict left-to-right order would dictate. These structural elements provide crucial context for generating subsequent code and can be predicted with high confidence from the prompt alone. The model exploits this hierarchical dependency structure to achieve greater parallelism: by unmasking the skeleton of the code early, it creates multiple independent sub-problems (filling in loop bodies, conditions, etc.) that can be decoded in parallel.

In contrast, mathematical reasoning in \Cref{fig:math_order} follows a smoother, more sequential progression. While there are minor deviations---such as unmasking numbers or operators that can be inferred from the problem statement---the overall pattern adheres closely to autoregressive order. This reflects the fundamentally sequential nature of step-by-step reasoning: each calculation step typically depends on previous results, limiting opportunities for parallel decoding. 

Overall, these unmasking patterns also validate our motivation: the learned planner creates masking distributions that locally resemble the random masking strategy from our case study (\Cref{fig:case_study}), selectively deviating from autoregressive order to unmask conditionally independent tokens in parallel while maintaining quality. The degree of non-autoregressive behavior adapts to task structure, with code exploiting more parallelism opportunities than sequential mathematical reasoning. Since consecutive tokens are usually highly dependent on each other in the true data distribution, the model's ability to generate many consecutive tokens in parallel demonstrates the mode filtering behavior discussed in the introduction: parallel generation of dependent tokens independently can be realized by filtering the modes of the joint distribution, trading diversity for speed. 

\section{Related Work}
\textbf{Reinforcement Learning for DLMs. } 
Recent work has explored post-training diffusion language models (dLLMs) with reinforcement learning to improve their reasoning and generation capabilities. Early examples include \texttt{d1}, which adapts GRPO-style optimization to the (approximate) likelihood structure of dLLMs \citep{zhao2025d1}, and DiffuCoder, which targets code generation for LLaDA-style masked diffusion models \citep{gong2025diffucoder}. A growing line of follow-up work focuses on improving the stability and bias/variance trade-offs of policy-gradient estimators for dLLMs \citep{tang_wd1_2025,rojas2025improving,lin2025boundary, wang2025spg, zhu2025enhancing, mmada}. Notably, \citet{wang2025revolutionizing,zhao2025diffpo} try to obtain more accurate likelihood by using more intermediate decoding steps. While these methods focus on reasoning capabilities, we propose to learn the optimal unmasking order and enable high quality mode filtering via RL. Close to our work is DCOLT \citep{huang2025dcolt}, which also learns an unmasking order policy with RL. However, their unmasking policy does not enable parallel decoding due to the adoption of the Plackett-Luce model. \citet{zhao2025diffpo} learns adaptive confidence thresholds for unmasking with RL, but their method still relies on logit-based heuristics for token selection rather than learning a flexible unmasking policy as in our work. Concurrent work \citet{jazbec2025} also learns an unmasking policy head with GRPO, but there are several important distinctions. First of all, they learn the unmasking policy from the logits rather than the last layer representation. Secondly, they do not jointly train the base model and the unmasking policy, but rather keep the base model fixed. Finally, they do not introduce an on-policy distillation reward to encourage mode-seeking behavior. As a result, they report only marginal benefits over logit-based heuristics, while we achieve better performance in both accuracy and NFE even when compared to SOTA distillation-based methods.

\textbf{DLM Acceleration.}
Most existing acceleration methods for masked dLLMs are either system level, such as enabling kv-cache and employing token dropping \citep{ma2025dkv,liu2025dllm,hu2025acceleratingdiffusionlanguagemodel,wu2025fastdllmtrainingfreeaccelerationdiffusion} or algorithmic level that designs better sampling strategies. Among the latter, there are two approaches: (i) training-free logit-based heuristic decoders \citep{wu2025fastdllmtrainingfreeaccelerationdiffusion,israel2025acceleratingdiffusionllmsadaptive,wei2025acceleratingdiffusionlargelanguage,ben-hamu_accelerated_2025} or (ii) \emph{distillation}-based approaches that fit the off-policy trajectories produced by a base model \citep{chen2025dparallellearnableparalleldecoding,d3llm}. Heuristic samplers often rely on confidence, entropy or other related uncertainty signals computed from the logits, and have recently enabled substantial inference speedups in dLLMs via parallel decoding and KV caching \citep{wu2025fastdllmtrainingfreeaccelerationdiffusion,hu2025acceleratingdiffusionlanguagemodel}. Distillation-based methods, such as pseudo-trajectory distillation in d3LLM \citep{d3llm}, learnable parallel decoding in dParallel \citep{chen2025dparallellearnableparalleldecoding}, and distribution-matching distillation in DiDi-Instruct \citep{zheng2025ultrafastlanguagegenerationdiscrete}, can further increase decoding efficiency by finetuning the base model to sample multiple steps of the base model in one step, which essentially filters the modes of the base model. These methods depend on the quality and coverage of the teacher trajectories, which also may become off-policy as training progresses. Our approach explicitly addresses this mismatch by using on-policy rollouts with GRPO, while also encouraging mode seeking behavior.

\textbf{Denoising Trajectory Planning. }
Beyond heuristic confidence-based planning, recent works have explored learned planning modules for MDLM trajectories. These approaches differ fundamentally in their planning strategies. P2 \citep{peng2025pathplanningmaskeddiffusion} introduces explicit planning and denoising sub-stages, enabling iterative refinement of existing unmasked tokens. DCOLT \citep{huang2025dcolt} employs the Plackett-Luce model to generate sequential unmasking orders optimized for accuracy rather than parallel decoding efficiency. DiFFPO \citep{zhao2025diffpo} learns adaptive per-prompt thresholds via off-policy RL but still selects tokens based on confidence heuristics. In the concurrent work, \citet{jazbec2025} learns an unmasking policy head with GRPO on token confidences from logits.

\section{Conclusion}

We present dUltra, a framework for learning optimal unmasking strategies to accelerate masked diffusion language models through reinforcement learning. Motivated by the need for \emph{mode filtering} and maximizing \emph{conditionally independent} tokens per step, we train an unmasking planner head jointly with the base model using GRPO with on-policy rollouts and distillation rewards. Empirically, dUltra outperforms distillation baselines (d3LLM, dParallel) in both accuracy and NFEs, validating the superiority of our method. We found that learned unmasking order adapts to the task structure naturally: code generation exploits more parallelism locally by learning an unmasking order resembling the random masking strategy while math reasoning decodes multiple consecutive tokens sequentially and in parallel, exhibiting mode filtering behaviors. This work demonstrates that learned unmasking policies guided by mode filtering and conditional independence principles enable progress towards "diffusion supremacy" over autoregressive LLMs.


\bibliographystyle{plainnat}
\bibliography{references}

\newpage
\appendix

\section{Experimental Setup}



\subsection{Implementation Details}
\subsubsection{Model Architecture} \label{app:arch}

Our model extends the LLaDA-Instruct model \citep{nie2025largelanguagediffusionmodels} with an unmasking planner head. The architecture consists of:

\textbf{Base Model.} We use LLaDA \citep{nie2025largelanguagediffusionmodels}, a masked diffusion language model with Llama-like architecture adapted for discrete diffusion. The model uses a special mask token and employs time-agnostic parametrization following recent findings that the optimal ELBO solution is independent of the time variable \citep{ou2025absorbingdiscretediffusionsecretly, sahoo_simple_2024}.

\textbf{Unmasking Planner Head.} Following the architecture in \citet{huang2025dcolt}, we introduce an unmasking planner head that takes the last-layer hidden states and outputs per-token unmasking probabilities:
\begin{equation}
p_{\text{unmask}}(x) = \sigmoid(\text{head}(h_x))
\end{equation}
where $h_x$ is the hidden state for token position $x$. The head consists of:
\begin{itemize}
    \item $L=1$ transformer blocks (LLaDABlock layers)
    \item Adaptive Layer Normalization (AdaLN) with timestep conditioning
    \item Mask embeddings
    \item Final linear projection to scalar logit
\end{itemize}
Since the unmasking planner head is small (a single transformer block), it adds only a small computational overhead relative to the base model.

\subsubsection{Training Configuration:}
\begin{itemize}
    \item Unmasking planner head finetuning: Learning rate 1e-6, batch size 4, cosine schedule with warmup
    \item GRPO: Block length 32/128, temperature 0.1, advantage clipping, group size 12, learning rate 5e-6
    \item Optimizer: AdamW with weight decay 0.01
    \item Hardware: Single H200 GPU
\end{itemize}

\subsection{Reward Function} \label{app:reward}

Our training uses a multi-component reward function that combines task-specific correctness verification, format compliance, and efficiency incentives. The total reward is computed as a weighted sum:
\begin{equation}
r_{\text{total}} = \sum_{i} \lambda_i \cdot r_i
\end{equation}
where $r_i$ are individual reward components and $\lambda_i$ are their weights. Below we detail the default configuration for mathematical reasoning and code generation tasks.

\subsubsection{Mathematical Reasoning Tasks (GSM8K, MATH500)}

\textbf{Correctness Reward} ($\lambda = 1.0$): Extracts the final answer from LaTeX $\backslash$boxed\{\} notation and verifies mathematical equivalence against ground truth using \texttt{math\_verify} library for symbolic comparison. Returns 2.0 for correct answers, 0.0 otherwise.

\textbf{Format Reward} ($\lambda = 1.0$): Rewards proper use of LaTeX boxed notation ($\backslash$boxed\{answer\}) for presenting the final answer. Returns 0.5 when answers are properly boxed, 0.0 otherwise. This encourages the model to follow standard mathematical writing conventions.

\textbf{Efficiency Reward} ($\lambda = 1.0$): Incentivizes faster sampling by penalizing the number of denoising steps: $r_{\text{step}} = -N_{\text{steps}} / 50$, where $N_{\text{steps}}$ is the total number of denoising iterations.

\textbf{On-Policy Distillation Reward} ($\lambda = 1.0$): Computes the average per-token log-likelihood of the generated completion under \texttt{Qwen2.5-Math-7B-Instruct} teacher model: $r_{\text{distill}} = \frac{1}{L}\sum_{i=1}^{L} \log p_{\text{teacher}}(c_i \mid \bm{q}, \bm{c}_{<i})$, where $L$ is the completion length. This encourages mode-seeking behavior by rewarding completions that the teacher assigns high probability.

\subsubsection{Code Generation Tasks (APPS, HumanEval, MBPP)}

\textbf{Correctness Reward} ($\lambda = 3.0$): Executes the generated code against test cases with a 10-second timeout per test. Returns 1.0 if all tests pass, 0.0 for any wrong outputs, and -0.05 for runtime errors (excluding syntax/indentation errors which return 0.0). Code is extracted from markdown fences and wrapped to support LeetCode-style class solutions.

\textbf{Code Fence Format Reward} ($\lambda = 1.0$): Encourages proper markdown code formatting with incremental scoring: +0.25 for any fence, +0.25 for \texttt{```python} fence, +0.25 for closing fence, +0.25 for no trailing text. Small penalty (0.001 per character) for extra content after closing fence. Maximum score: 1.0.

\textbf{Efficiency Reward} ($\lambda = 0.1$): Same as math tasks.

\textbf{On-Policy Distillation Reward} ($\lambda = 1.0$): Uses \texttt{Qwen2.5-Coder-7B-Instruct} teacher model. Computation is identical to math tasks but with domain-specific teacher.

\newpage
\section{Training Algorithm}
\label{app:alg}
\begin{algorithm}[H]
\caption{Accelerating diffusion LLM with unmasking order planning using GRPO}
\KwIn{diffusion LLM model $\theta$, unmasking planner head $\phi$, dataset $\mathcal{D}$, reward\_func, advantage clipping threshold $C$}
\While{$\theta$ and $\phi$ not converged and maximum epochs not reached}{
    Sample questions $q \sim \mathcal{D}$\\
    \For{$g \gets 1$ \KwTo $G$}{ \tcp{Generate a group of $G$ trajectories}
        Initialize $x^g$ with $q$ and mask tokens\\
        $k \gets 0$ \tcp{number of steps}
        \While{$x^g$ has mask tokens}{
            Run the model forward once to get the logits $l$ and last-layer hidden state $h$\\
            Calculate unmasking probability $\pi_{\text{unmask}} = \sigmoid(\phi(h))$ for each mask token \\
            Calculate token probability $\pi_{\text{token}} = \softmax(l)$\\
            Sample the mask tokens to unmask $\posk \sim \text{MultivariateBernoulli}(\pi_{\text{unmask}})$\\
            Sample the value of unmasked tokens $\m{x}_{\posk}^g \sim \pi_{\text{token}}$\\
            $k\gets k + 1$\\
        }
        $k^g \gets k$\\
        $r^g \gets \text{reward\_func}(x^g)$ \tcp{Compute the rewards}
    }
    \For{$g \gets 1$ \KwTo $G$}{
        $A^g \gets r^g - \text{mean}(r^{1:G})$ \tcp{Compute advantages} 
        \If{$A^g < C$}{
            $A^g \gets 0$ \tcp{Advantage clipping}
        } 
    }
    $N \gets \max(k^g)$\\
    \tcp{Loop through the group}
    \For{$g \gets 1$ \KwTo $G$}{
        \tcp{Loop through denoising steps}
        \For{$n \gets 1$ \KwTo $N$}{ 
            \If{$k^g> n$}{
                \tcp{Compute $\pi_\theta$ and losses}
                $\pi\big(\bm{x}^{g,n} \mid \bm{x}^{g,n-1}\big)
= \pi_{\text{select}}\big(\textbf{Pos}^{(n)} \mid \bm{x}^{g,n-1}\big)
\cdot \pi_{\text{token}}\big(\bm{x}^{g,n} \mid  \bm{x}^{g,n-1}, \textbf{Pos}^{(n)}\big)$ \tcp{\Cref{eq:likelihood}}
                $\mathcal{L}_{\theta,n} \gets -\frac{1}{\norm{\bm{x}}\cdot G} \sum_{g=1}^G
                \frac{\pi(\bm{x}^{g,n}|\bm{x}^{g,n-1})}{\sg(\pi(\bm{x}^{g,n}|\bm{x}^{g,n-1}))} A^g$ \tcp{sg is stop gradient}
                Accumulate gradient $\nabla_\theta \mathcal{L}_{\theta,n}$\\
            }
        }
    }
    Update $\theta$ with accumulated gradients $\sum_{n=1}^N \nabla_\theta \mathcal{L}_{\theta,n}$\\
    Zero out accumulated gradients\\
}
\end{algorithm}

\section{Quality-speed trade-off with multiplicative factors} \label{app:trade-off}
\begin{figure}[h]
\centering
\includegraphics[width=\textwidth]{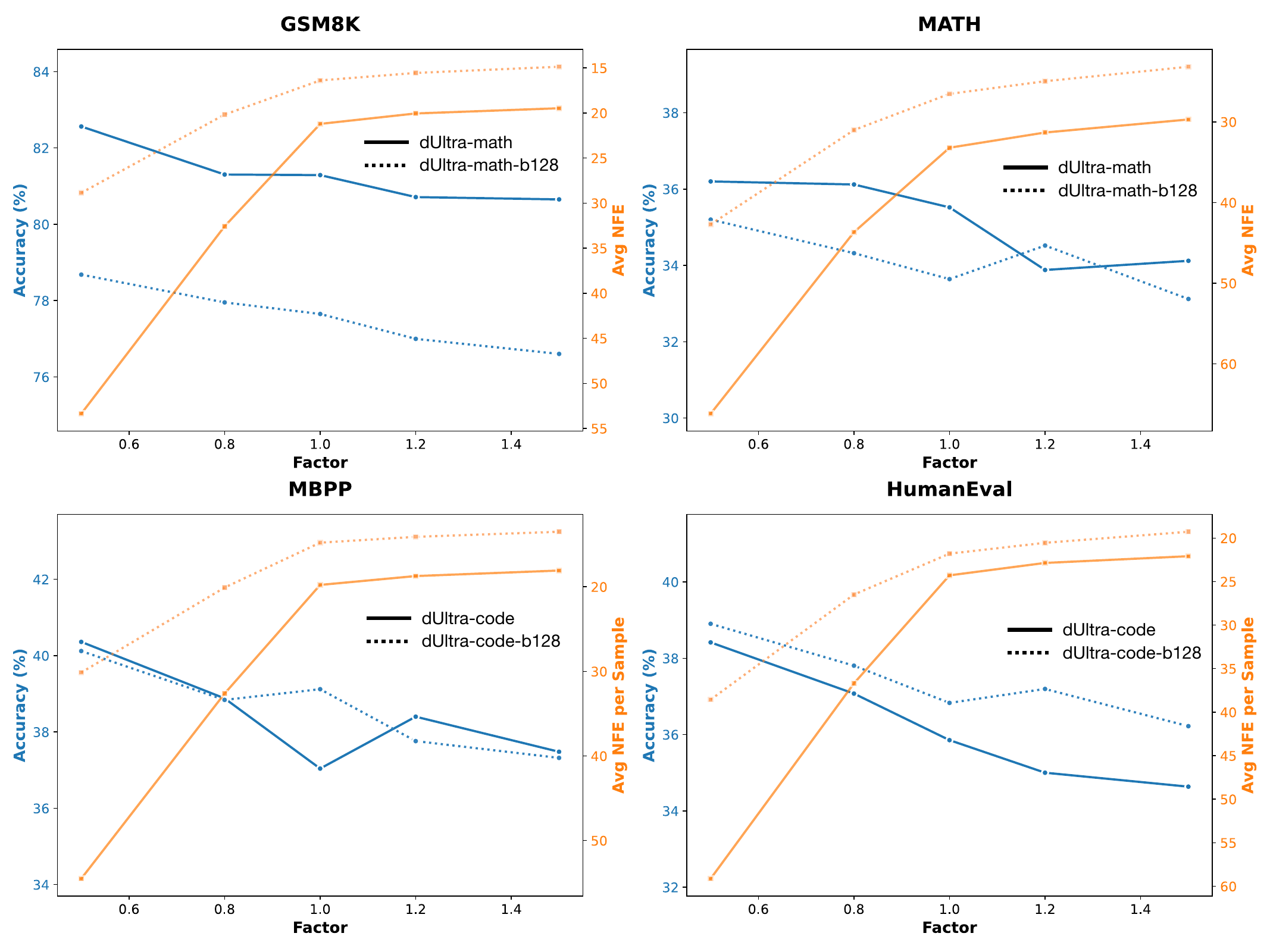}
\caption{Accuracy vs. NFE trade-off on GSM8K, MATH500, HumanEval, MBPP. We vary the multiplicative factor (x axis) applied to the predicted unmasking probability. }
\label{fig:pareto_factor}
\end{figure}

The unmasking probability predicted by the unmasking planner head directly controls the expected number of tokens unmasked at each denoising step. This allows us to strike a balance between sampling efficiency (NFE) and task performance (accuracy). By multiplying the unmasking probability by a factor $\alpha$ \footnote{For \Cref{tab:b32_results} and \Cref{tab:b128_results}, we use the unscaled unmasking probability, i.e. $\alp=1$.}, we are able to adjust the sampling speed at inference time without retraining the model. \Cref{fig:pareto_factor} illustrates the accuracy vs. NFE trade-off on all tasks for both block size 32 and block size 128.

\begin{figure}[h]
\centering
\includegraphics[width=\textwidth]{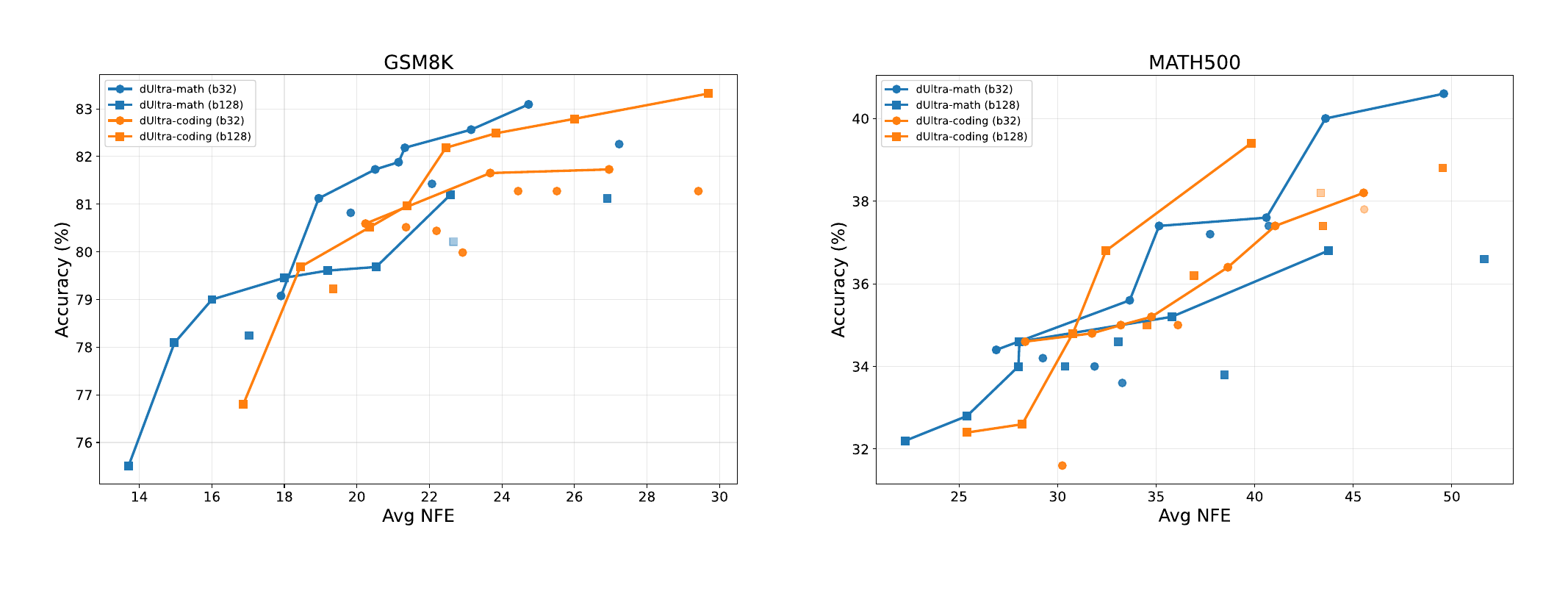}
\caption{Pareto frontier of dUltra-math vs dUltra-coding on math reasoning tasks.}
\label{fig:pareto_mathcoding}
\end{figure}
\end{document}